\documentclass[conference]{IEEEtran}
\IEEEoverridecommandlockouts
\usepackage{amsmath,amssymb,amsfonts}
\usepackage{algorithmic}
\usepackage{graphicx}
\usepackage{textcomp}
\usepackage{xcolor}

\definecolor{myblue}{rgb}{0,0,0.6}
\usepackage[colorlinks=true, citecolor=myblue, linkcolor=myblue, urlcolor=myblue]{hyperref}
\usepackage{url}
\usepackage{graphicx}
\usepackage{comment}
\usepackage{ulem}
\usepackage{bm}
\usepackage{amsmath, amssymb}
\usepackage{type1cm}
\usepackage{multirow}
\usepackage{algorithm2e}
\usepackage{wrapfig}
\usepackage{subfig}
\usepackage[numbers]{natbib}

\def\BibTeX{{\rm B\kern-.05em{\sc i\kern-.025em b}\kern-.08em
    T\kern-.1667em\lower.7ex\hbox{E}\kern-.125emX}}
\begin{document}

\title{Model-Based Counterfactual Explanations Incorporating Feature Space Attributes\\ for Tabular Data
}

\author{\IEEEauthorblockN{Yuta Sumiya}
\IEEEauthorblockA{University of Electro-Communications \\
Chofu, Tokyo 182-8585, Japan \\
Email: sumiya@uec.ac.jp}
\and
\IEEEauthorblockN{Hayaru Shouno}
\IEEEauthorblockA{University of Electro-Communications \\
Chofu, Tokyo 182-8585, Japan \\
Email: shouno@uec.ac.jp}}

\maketitle

\begin{abstract}
Machine-learning models, which are known to accurately predict patterns from large datasets, are crucial in decision making. Consequently, counterfactual explanations—methods explaining predictions by introducing input perturbations—have become prominent. These perturbations often suggest ways to alter the predictions, leading to actionable recommendations. However, the current techniques require resolving the optimization problems for each input change, rendering them computationally expensive. In addition, traditional encoding methods inadequately address the perturbations of categorical variables in tabular data. Thus, this study propose FastDCFlow, an efficient counterfactual explanation method using normalizing flows. The proposed method captures complex data distributions, learns meaningful latent spaces that retain proximity, and improves predictions. For categorical variables, we employed TargetEncoding, which respects ordinal relationships and includes perturbation costs. The proposed method outperformed existing methods in multiple metrics, striking a balance between trade offs for counterfactual explanations. The source code is available in the following repository: \url{https://github.com/sumugit/FastDCFlow}.

\end{abstract}


\section{Introduction}
\label{sec:introduction}
\begin{figure}[t]
  \begin{center}
  \includegraphics[width=1.0\columnwidth]{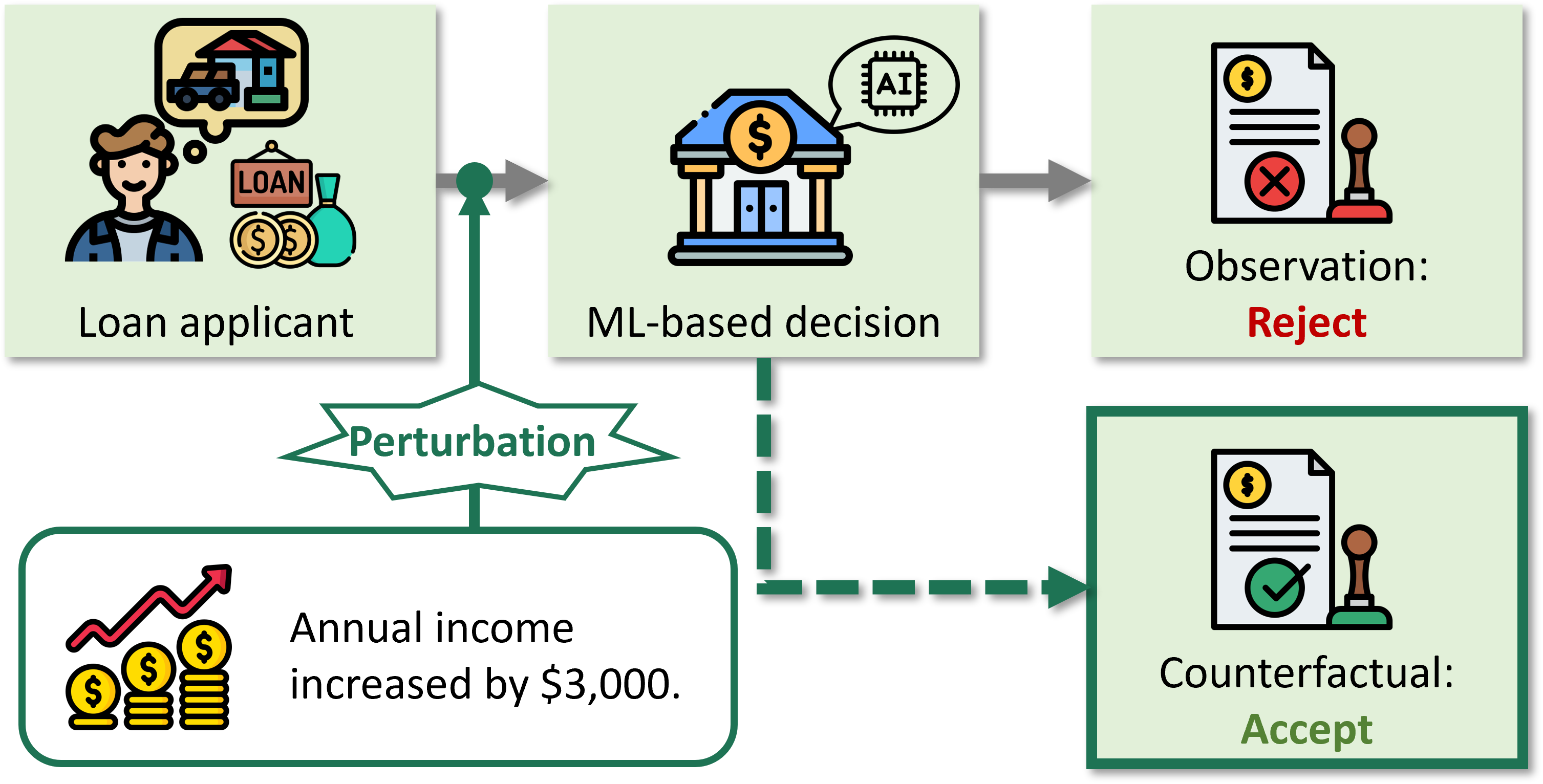}
  \caption{Counterfactual scenario.}
  \label{fig:CE}
  \end{center}
\end{figure}

Machine learning (ML) technologies seek to replicate the learning capabilities of computers. With recent advancements, ML can now address decision-making tasks that were traditionally determined by humans. Such decisions often require counterfactual explanations (CE). Counterfactuals envision unobserved hypothetical scenarios. For instance, as shown in Figure \ref{fig:CE}, if a bank's algorithm denies a loan, a counterfactual might reveal that an extra \$3,000 in annual income would have secured approval, guiding the applicant towards future success.

When implementing CE in ML, methods often introduce input perturbations to optimize the target variable predictions. These altered inputs produce counterfactual samples (CF), representing unobserved scenarios. As outlined by \cite{wachter2017counterfactual}, CFs should satisfy two constraints that are often in a trade-off relationship: validity and proximity. Validity ensures that perturbations modify inputs such that ML models yield the desired output, such as changing loan denial to approval in two-class classifications. Whereas, proximity requires the perturbations to remain as close as possible to the original input. However, relying solely on these two metrics can be misleading, as models may appear to perform well while generating nearly identical CFs. Given that they are not necessarily unique, assessing the diversity among CFs is also crucial. 
Interestingly, as reported by \cite{pawelczyk2022exploring}, CF generation echoes the principles of adversarial attacks \citep{szegedy2013intriguing, ballet2019imperceptible}, wherein inputs are perturbed until the predicted class shifts, typically when the binary prediction probability reaches $\geq$ 0.5. However, in contrast to adversarial attacks, which focus on threshold shifts, we advocated CFs that span various prediction probabilities. 

The primary goal of CE is to generate a broad range of diverse CFs that adhere to both validity and proximity. However, there are two challenges with the current methods. The first is the presence of both categorical and continuous variables in the tabular data. For instance, in the context of a loan application, suggesting a counterfactual transition from a ``high school graduate" to a ``PhD holder" may be impractical. If a counterfactual suggests upgrading to ``college graduate," this could be a more realistic change for approval. Current preprocessing methods, such as OneHotEncoding (OHE) and LabelEncoding (LE), as in \cite{mothilal2020explaining, duong2023ceflow}, tend to produce unrealistic outcomes. OHE often misinterprets categorical shifts, equating a jump from high school to PhD with a change to college. With LE, perturbations become non differentiable, which is unsuitable for deep-learning models. The second challenge is related to the overhead of crafting individual CFs for each input. As described in \cite{rodriguez2021beyond}, generating CFs often involves the optimization of variables tailored to each input, which results in inefficiencies as the volume of inputs expands and impacts storage and execution time.

To address these challenges, we introduced fast and diverse CEs using normalizing flows (FastDCFlow). Within FastDCFlow, we managed the perturbations of categorical variables using TargetEncoding (TE). This method aggregated the average value of the target variable for each categorical level using training data and substituted the categorical values with their corresponding averages. Thus, categorical variables were effectively transitioned into continuous variables, maintaining a meaningful order of magnitude relative to the predicted values. Furthermore, to generate optimal CFs, we leveraged normalizing flows \citep{dinh2014nice, dinh2016density, kingma2018glow} to capture complex data distributions through a reversible latent space. In summary, our main contributions include: (1) Using TE, we lower the model's learning cost and diversify the generated CFs' predicted values. (2) We introduce FastDCFlow, a method that learns a latent space for ideal CF generation using normalizing flows. We evaluate the CFs against multiple metrics and compare with other methods to confirm our approach's effectiveness.

\section{Related work}
\label{sec:related_work}
In this study, we broadly classified the generation algorithms of existing research into two categories: input- and model-based methods.\\

\noindent
\textbf{Input-based methods.} These methods determine the optimal set of CFs for each unobserved input (test input), requiring relearning of variables with every input alteration. Reference \cite{wachter2017counterfactual} optimized variables for a single CF, employing a gradient descent-based algorithm. Reference \cite{mothilal2020explaining} introduced DiCE, which was designed to produce multiple CFs. DiCE optimizes sets of multiple CFs by adding an explicit diversity term based on a determinant from the distance matrix between variables. However, the computational complexity increases with variable dimensionality and the number of CFs produced. In contrast to gradient-based methods, a genetic algorithm (GA) was used for efficient generation with increasing CF count. Reference \cite{schleich2021geco} proposed GeCo, in which CFs as were conceptualized as genes and cross-over and mutation were executed across generations. Reference \cite{dandl2020multi} proposed MOC. They framed objective functions through multi-objective optimization capturing trade-offs. Critically, these methods independently optimize the dimentions of the observed variable, overlooking the impact of unobserved common factors in the data generation process.

When considering the influence of unobserved common factors, one approach perturbs the latent variables presumed to underpin the observed ones. Reference \cite{rodriguez2021beyond} introduced DIVE, which employed the variational auto-encoder (VAE) \citep{kingma2013auto, rezende2014stochastic, mnih2014neural, gregor2014deep} to determine the optimal perturbations for latent variables. Although DIVE delivers insightful CEs in the image domain, its efficacy on tabular data with categorical variables remains unverified. Many studies have transformed categorical variables into continuous variables using OHE or LE. However, these transformations lack a magnitude relationship between the values. Taking into account dimension expansion and sparsity issues, these methods are unsuitable for CE, where variable perturbations are paramount. Reference \cite{duong2023ceflow} presented CeFlow, which incorporated normalizing flows into tabular data to learn an added perturbation. However, as this perturbation is designed to generate only one CF for each test input, it fails to ensure diversity when creating multiple CFs. Additionally, CeFlow, employing variational dequantization as per \citep{hoogeboom2021learning}, faces challenges due to the lack of explicit ordering in categorical variables. Consequently, the continuous variables lack proximity to their inherent values.\\

\noindent
\textbf{Model-based methods.} These approaches involve direct learning of the CF generation model using training data. They requires only one learning session. After stabilizing the model with trained parameters, CFs can be readily produced for test input. Reference \cite{looveren2021interpretable} employed a spatial partitioning data structure, the k-d tree \citep{bentley1975multidimensional}, to create a set of training data points as CFs close to the test input within the target prediction class. However, this search was limited to the scope of observed data points, which rendered the generating of CFs for data not previously observed challenging. Reference \cite{mahajan2019preserving} introduced a technique that directly learned the latent space for CF generation using VAE. In contrast to VAE as an input-based method that optimized perturbations, this method offered the benefit of eliminating the need for model retraining for each input. However, because VAE assumes a latent distribution based on continuous variables, certain issues remain in the processing of data that include categorical variables.

Input-based methods require optimization problems to be solved for each input. This process becomes increasingly inefficient as the number of verification inputs increases. Therefore, the proposed FastDCFlow aimed to produce CFs quickly, adhering to the necessary constraints for tabular data. To achieve this, a model-based approach was employed.

\section{Proposed method}
\label{sec:Proposed_method}
\begin{figure}[t]
  \begin{center}
  \includegraphics[width=1.0\columnwidth]{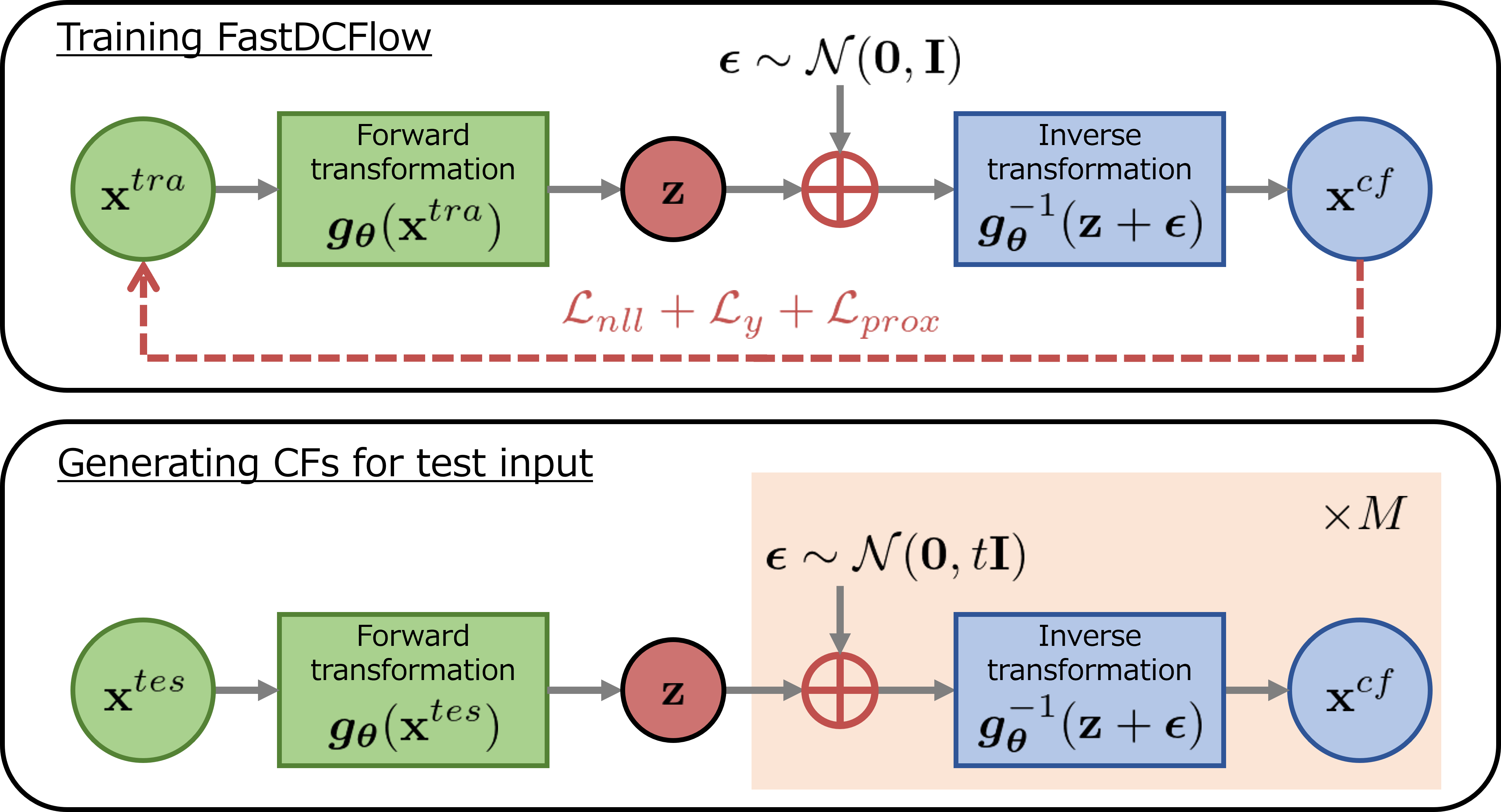}
  \caption{Overview of CFs training and generation using FastDCFlow.}
  \label{fig:FastDCFlow}
  \end{center}
\end{figure}
Consider an input space $\mathcal{X}$ with input $\mathbf{x} \in \mathcal{X}$ and its corresponding target $y \in \{0, 1\}$. The ML model $f(\mathbf{x})$ predicts the probability of a positive outcome, where $f(\mathbf{x}) = \hat{y}$ and $\hat{y} \in [0, 1]$. It is assumed that all categorical variables are transformed into continuous variables using TE. The purpose of CE is to generate perturbed inputs $\mathbf{x}^{cf}$ that should be $f(\mathbf{x}^{cf}) > f(\mathbf{x})$ for the observed input $\mathbf{x}$. To support this, the latent representation $\mathbf{z} \in \mathcal{Z}$ of the input $\mathbf{x}$ was trained. Ideally, $\mathbf{z}$ captures the representation of the unobserved common factors, ensuring both proximity and diversity in the input space.

\subsection{FastDCFlow}
We employed normalizing flows to prioritize efficient generation and precise likelihood computation. These flows are generative models that use a function $\bm{g}_{\bm{\theta}}$, parameterized by $\bm{\theta}$, which transforms the distribution of a dataset into another distribution in the latent space. In the forward transformation, latent variables $\mathbf{z}\sim p_{\mathcal{Z}}$ are derived by mapping $\bm{g}_{\bm{\theta}}: \mathcal{X}\rightarrow \mathcal{Z}$, transitioning from the input space $\mathcal{X}$ to the latent space $\mathcal{Z}$. The distribution of the dataset is transformed back using the inverse $\bm{g}_{\bm{\theta}}^{-1}: \mathcal{Z}\rightarrow \mathcal{X}$, resulting in a random variable $\mathbf{x}\sim p_{\mathcal{X}}$. The transformation of the probability density function for $\mathbf{x} = \bm{g}_{\bm{\theta}}^{-1}(\mathbf{z})$ is expressed as:
\begin{align}
    \begin{split}
    \log{\left(p_{\mathcal{X}}(\mathbf{x})\right)} = & \log{\left(p_{\mathcal{Z}}(\bm{g}_{\bm{\theta}}(\mathbf{x}))\right)} \\
    & + \log{\left(\left|\text{det}\left(\frac{\partial \bm{g}_{\bm{\theta}}(\mathbf{x})}{\partial \mathbf{x}}\right)\right|\right)}.          
    \end{split}  
\end{align}
For effective CF generation, it is essential to produce samples quickly and diversely to enhance the input predictions while preserving proximity. In our proposed FastDCFlow method, beyond precise likelihood computation, we incorporated additional losses, accounting for the constraints that CFs must satisfy. Furthermore, to accelerate the sample generation, we implemented a model-based method that learned the latent space to enable CF generation in a single training session.

\subsection{Training and generation}
Figure \ref{fig:FastDCFlow} illustrates the comprehensive framework of the proposed method. Initially, we investigated the FastDCFlow training process. Subsequently, we explained the approach to foster diversity during test input considerations.\\

\noindent
\textbf{Minimization of the negative log-likelihood.} The number of training inputs is represented as $N^{tra}$: Each input pair is denoted by $\mathcal{D} = \{\mathbf{x}_i^{tra}\}_{i=1}^{N^{tra}}$. Owing to the property of normalization flows, which enables the direct evaluation of parameter likelihoods, the negative log-likelihood (NLL) was integrated into the loss function:
\begin{align}
    \begin{split}
    \mathcal{L}_{nll}(\bm{\theta}, \mathcal{D}) = & -\sum_{i=1}^{N^{tra}}\bigg(\log{\left(p_{\mathcal{Z}}(\bm{g}_{\bm{\theta}}(\mathbf{x}_{i}^{tra}))\right)} \\
    & + \log{\left(\left|\text{det}\left(\frac{\partial \bm{g}_{\bm{\theta}}(\mathbf{x}_{i}^{tra})}{\partial \mathbf{x}_{i}^{tra}}\right)\right|\right)}\bigg).
    \end{split}
\end{align}
\noindent
\textbf{Validity and proximity.} In the process of generating CFs, the initial step involves mapping the input $\mathbf{x}^{tra}$ onto the latent space through $\mathbf{z} = \bm{g}_{\theta}(\mathbf{x}^{tra})$. To prioritize sampling in the neighborhood of the input, a perturbed latent variable $\mathbf{z}^{*}$ is derived by introducing a reparametrization trick \citep{kingma2013auto} that adds Gaussian noise $\bm{\epsilon}$ from the distribution $\mathcal{N}(\bm{0}, \bm{I})$, resulting in $\mathbf{z}^{*} = \mathbf{z} + \bm{\epsilon}$. Although this approach is simple, it allows efficient sampling in various directions within the latent space and simultaneously preserves proximity to the input. $\mathbf{x}^{cf}$ is generated from this perturbed variable as $\mathbf{x}^{cf} = \bm{g}_{\theta}^{-1}(\mathbf{z}^{*})$:

CFs are fundamentally designed to satisfy two primary criteria: validity and proximity to the input. The validity of our model was measured using the cross-entropy loss function, defined as
\begin{align}
    \begin{split}
    \mathcal{L}_y(\bm{\theta}, \mathcal{D}) = &-\sum_{i=1}^{N^{tra}}\left(y_i^{cf}\log{(f(\mathbf{x}_{i}^{cf}))} \right. \\
    & \left. + (1-y_i^{cf})\log(1-f(\mathbf{x}_{i}^{cf}))\right)\\
    & = -\sum_{i=1}^{N^{tra}}\log{(f(\mathbf{x}_{i}^{cf}))},
    \end{split}
\end{align}
where $y_i^{cf}$ is the anticipated prediction value for CF. For binary classification tasks, the positive class probability is maximized by setting $y_i^{cf}=1$.

Considering the transformation of continuous spectrum, we use the weighted square error to quantify proximity, which includes a hyperparameter $\bm{w}$ to adjust the perturbability of the features.
\begin{equation}
    \mathcal{L}_{wprox}(\bm{\theta}, \mathcal{D}) = \sum_{i=1}^{N^{tra}}\| \bm{w}*(\mathbf{x}_{i}^{tra} - \mathbf{x}_{i}^{cf}) \|_2^{2},
\end{equation}
where $\ast$ is the element product of the vectors.\\

\noindent
\textbf{Optimization.}
The overall optimization problem can be represented as
\begin{equation}
    \bm{\theta}^{*} = \underset{\bm{\theta}}{\operatorname{argmin}}\ \lambda\mathcal{L}_{nll}(\bm{\theta}, \mathcal{D}) + \mathcal{L}_{y}(\bm{\theta}, \mathcal{D}) + \mathcal{L}_{wprox}(\bm{\theta}, \mathcal{D}).
\end{equation}
In this equation, the hyperparameter $\lambda$ balances the importance of NLL, validity, and proximity. 

Once FastDCFlow's parameters $\bm{\theta}$ are trained, the model can be fixed using these parameters. This approach accelerates the generation of a CF set for any test input $\mathbf{x}^{tes}$, thereby rendering it more efficient than input-based methods. To adjust diversity during generation without adding any explicit diversity loss term, we introduced a temperature parameter $t$ and added Gaussian noise to the latent space with $\bm{\epsilon}_t \sim \mathcal{N}(\bm{0}, t\bm{I})$. This simple revision boosts model uncertainty, which is effective for sampling a more diverse range of CFs from the latent space.

Given $N^{tes}$ as the number of test inputs, with $\mathcal{T}=\{\mathbf{x}_i^{tes}\}_{i=1}^{N^{tes}}$ representing each input pair, and $M$ as the number of CFs generated for each input, this can be carried out in parallel. a comprehensive procedure, including training, is presented in Algorithm\ref{alg:1}.

In the TE transformation, two functions are utilized: $\text{fit\_transformTE}(\bullet)$, which substitutes categorical variables in training data with target means via out-of-fold calculations, and $\text{transformTE}(\bullet)$, which is the same for all training data. The CF sets derived from the test inputs were reverted to their initial categorical variables using the $\text{inverse\_transformTE}(\bullet)$ function. During this step, a binary search tree assisted in mapping the variables to the closest categorical variable level.

\section{Evaluation}
\label{sec:evaluation}
\begin{table}[t]
    \centering
    \caption{Evaluation metrics. $I(\bullet)$ is the indicator function and the expression $\binom{n}{2}$ represents the number of combinations, denoted as ${}_n \mathrm{C}_k$.}
    \begin{tabular}{l|l}
    \hline
    \multicolumn{1}{c|}{Metrics} & \multicolumn{1}{|c}{Definition} \\
    \hline
    Inner diversity & $\text{ID} = -\frac{1}{N^{tes} \binom{M}{2}}\sum_{i=1}^{N^{tes}}\sum_{j\neq k}^{M} \frac{{\mathbf{x}_{ij}^{cf}}^{\top}\mathbf{x}_{ik}^{cf}}{\|\mathbf{x}_{ij}^{cf}\| \|\mathbf{x}_{ik}\|}$ \\[1.0em]
    Outer diversity & $\text{OD} = -\frac{1}{\binom{N^{tes}}{2}} \sum_{i\neq j}^{N^{tes}} \frac{\overline{\mathbf{x}^{cf}_i}^{\top} \overline{\mathbf{x}^{cf}_j}}{\|\overline{\mathbf{x}^{cf}_i}\| \|\overline{\mathbf{x}^{cf}_j}\|}$ \\[1.0em]
    Proximity & $\mathrm{P} = -\frac{1}{N^{tes}}\sum_{i}\frac{1}{M}\sum_{j}\|\mathbf{x}_{i}^{tes}-\mathbf{x}_{ij}^{cf}\|_2$ \\[1.0em]
    Validity & $\text{V} = \frac{1}{N^{tes}}\sum_{i}\frac{1}{M}\sum_{j}I\left(f(\mathbf{x}_{ij}^{cf})-f(\mathbf{x}_{i}^{tes}) > 0\right)$ \\[1.0em]
    Run time & $\text{RT} = \frac{1}{N^{tes}}T(N^{tes}, M)$ \\
    \hline
    \end{tabular}
    \label{tab:metrics}
\end{table}
\RestyleAlgo{ruled}
\begin{algorithm}[t]
\caption{Generating CFs}
\label{alg:1}
\KwData{Training inputs $\mathcal{D} = \{\mathbf{x}_i^{tra}\}_{i=1}^{N^{tra}}$, test inputs $\mathcal{T} = \{\mathbf{x}_i^{tes}\}_{i=1}^{N^{tes}}$, pretrained binary probability prediction model $f(\bullet)$.}
\nl  Apply TE to training and test inputs to convert categorical variables into continuous variables:
\begin{align*}
    \mathcal{D} &\leftarrow \text{fit\_transformTE}(\mathcal{D}),\\
    \mathcal{T} &\leftarrow \text{transformTE}(\mathcal{T}).
\end{align*}
\nl Learn the parameters $\bm{\theta}^{*}$ of the FastDCFlow function $\bm{g}_{\bm{\theta}}$ from NLL, validity, and proximity losses:
\begin{equation}
    \bm{\theta}^{*} = \underset{\bm{\theta}}{\operatorname{argmin}}\ \lambda\mathcal{L}_{nll}(\bm{\theta}, \mathcal{D}) + \mathcal{L}_{y}(\bm{\theta}, \mathcal{D}) + \mathcal{L}_{wprox}(\bm{\theta}, \mathcal{D}). \nonumber
\end{equation}
\nl \For{$i=1, \cdots, N^{tes}$}{
    \nl Obtain latent variable $\mathbf{z}_i$ using forward transformation, $\mathbf{z}_i = \bm{g}_{\bm{\theta}}(\mathbf{x}_i^{tes})$, $\mathbf{z}_i\sim p_{\mathcal{Z}}$. \\
    \nl \For{$j=1, \cdots, M$}{
        \nl Obtain perturbed latent variable $\mathbf{z}_{ij}^{*} = \mathbf{z}_{i} + \bm{\epsilon}_{tj}$ with $\bm{\epsilon}_{tj} \sim \mathcal{N}(\bm{0}, t\bm{I})$.\\
        \nl Generate counterfactual sample $\mathbf{x}_{ij}^{cf} = \bm{g}_{\theta}^{-1}(\mathbf{z}_{ij}^{*})$, $\mathbf{x}_{ij}^{cf}\sim p_{\mathcal{X}}$.
    }
    \nl Reverse-transform categorical variables to their original levels:
    \begin{equation}
        \mathbf{x}_{ij}^{cf} \leftarrow \text{inverse\_transformTE}(\mathbf{x}_{ij}^{cf}). \nonumber
    \end{equation}
}
\KwResult{Set of CFs $\left\{\mathcal{S}_{i}\right\}_{i=1}^{N^{tes}}$, where $\mathcal{S}_{i} = \{\mathbf{x}_{ij}^{cf}\}_{j=1}^{M}$.}
\end{algorithm}
\noindent
\textbf{Validation of TE effectiveness.} 
We employed two distinct training datasets, one using OHE and the other using TE, to train both the ML model for probability estimation and FastDCFlow. By evaluating the classification performance difference on the test data using a t-test, we ensured consistent performance across both encoding techniques (OHE and TE).

We examined the CE trends produced by FastDCFlow. For the $i$-th test input, let the predicted value be denoted as $\hat{y}_i^{tes}=f(\mathbf{x}_{i}^{tes})$ and the average predicted value across all test inputs as $\hat{y}^{tes} = \frac{1}{N^{tes}}\sum_{i}\hat{y}_i^{tes}$.

For each test input, a set of $M$ CFs was generated and represented as $\mathcal{S}_{i} = \{\mathbf{x}_{ij}^{cf}\}_{j=1}^{M}$. The $j$-th predicted value within this set was $\hat{y}_{ij}^{cf}=f(\mathbf{x}_{ij}^{cf})$. The average predicted value across these $M$ CFs was $\hat{y}_i^{cf} = \frac{1}{M}\sum_{j}\hat{y}_{ij}^{cf}$, and the overall average predicted value for all CFs is expressed as $\hat{y}^{cf}=\frac{1}{N^{tes}}\sum_{i}\hat{y}_i^{cf}$.

These metrics aid in ascertaining the changes in the CF prediction values. To address instances wherein the CF prediction values were consistently the same irrespective of the test input, we also validated them based on the standard deviation of $\hat{y}_i^{cf}$. Considering that OHE and TE yield varying feature dimensions post-transformation, both the variability of $\hat{y}_i^{cf}$ for individual test inputs and the standard deviation values were compared.\\

\noindent
\textbf{Performance metrics for CE.} We introduce five specialized metrics to evaluate CE. Both FastDCFlow and its competitors employed TE-transformed datasets, treating all features as continuous variables. Table \ref{tab:metrics} lists the metrics.
\begin{itemize}
    \item \textbf{Inner diversity (ID)}: The diversity metric focuses on the cosine similarity within a set of $M$ CFs for a single test input. This metric emphasizes direction rather than $l_2$ distance to avoid dependency on the subsequently mentioned proximity metric.
    \item \textbf{Outer diversity (OD)}: The diversity metric focuses on the cosine similarity between CF pairs generated for different test inputs. The expression $\overline{\mathbf{x}^{cf}_i} = \frac{1}{M}\sum_{j}\mathbf{x}^{cf}_{ij}$ represents the average of a set of CFs for a single test input.
    \item \textbf{Proximity (P)}: Utilizing the $l_2$ norm, we measured how close each generated CF was to its corresponding test input.
    \item \textbf{Validity (V)}: We computed the mean improvement in the CFs' predicted outcomes ($y_{ij}^{cf}$) over the test inputs ($y_{i}^{tes}$).
    \item \textbf{Run time (RT)}: This metric considers the time component $T(N^{tes}, M)$, which measures the time required to generate $M$ CFs for each input of the test size $N^{tes}$.
\end{itemize}
Except for RT, the sign of the values has been adjusted so that larger values indicate higher performance. When the ID value reaches its minimum of -1.0, it indicates that all generated CFs are identical. In such cases, since only a single CF influences the results, comparisons of performance in OD, P, and V are not performed. For RT, a shorter generation time per CF indicates superior performance.

\section{Experiment}
\label{sec:experiment}
\subsection{Datasets and preprocessing}
\begin{table}[t]
\centering
\caption{Dataset composition.}
\begin{tabular}{c|c|cccc}
\hline
\multicolumn{2}{c|}{Dataset} & \# Rec & \# Cat & \# Con & \# Dim \\ \hline
\multirow{2}{*}{Adult} & OHE & \multirow{2}{*}{32,561} & \multirow{2}{*}{6} & \multirow{2}{*}{2} & 30 \\
& TE & & & & 8 \\ \hline
\multirow{2}{*}{Bank} & OHE & \multirow{2}{*}{11,162} & \multirow{2}{*}{9} & \multirow{2}{*}{7} & 52  \\
& TE & & & & 16 \\ \hline
\multirow{2}{*}{Churn} & OHE & \multirow{2}{*}{7,043} & \multirow{2}{*}{16} & \multirow{2}{*}{3} & 47 \\
& TE & & & & 19 \\ \hline
\end{tabular}
\label{tab:datasets}
\end{table}
To evaluate the proposed approach, we used three open datasets: Adult \citep{misc_adult_2}, Bank \citep{misc_bank_marketing_222}, and Churn \citep{macko_2019_telco}, which integrate both categorical and continuous variables. The Adult dataset, sourced from the US Census, focuses on factors influencing an individual's annual income, classifying whether it exceeds \$50k. The Bank dataset, which is from a Portuguese bank's marketing efforts, centers on whether customers opt for a term deposit. The Churn dataset from IBM presents fictional telecom customer churn indicators. Although the Adult dataset is distinct in its six categorical variables with a vast range of values, both the Bank and Churn datasets stand out for their high dimensionality, with the latter heavily leaning towards categorical data. Across all datasets, the task of the ML model was to predict the probability of a target variable. 

In our analysis, we prepared two versions of the datasets, transforming categorical variables using OHE and TE, and standardizing them so that they have a mean of 0 and a variance of 1. Table \ref{tab:datasets} shows the details of the datasets. \# Rec, \# Cat, \# Con, and \# Dim represent the number of records, count of categorical variables, count of continuous variables, and the dimension after encoding. The data is partitioned into a 90\% training set and a 10\% test set. The test inputs for CF generation are those with a predicted probability of less than 0.5 from the pretrained ML model. We implemented a 10-fold on the training data for TE.

\subsection{Baselines}
\noindent
\textbf{Existing methods.}
To assess the effectiveness of TE, we employed datasets preprocessed with OHE and TE, which were evaluated using standard deviations. Compared with other competing models, five gradient-based models were considered: DiCE \citep{mothilal2020explaining}, CeFlow \citep{duong2023ceflow}, CF\_VAE \citep{mahajan2019preserving}, CF\_CVAE, and CF\_DUVAE. As described in section \ref{sec:related_work}, DiCE and CeFlow are input-based methods for generating counterfactuals, whereas the other methods employ model-based methods. CF\_VAE uses VAE to generate counterfactuals. By contrast, CF\_CVAE extends CF\_VAE by incorporating a conditional variational autoencoder (CVAE) \citep{sohn2015learning} to handle category labels explicitly. Furthermore, CF\_DUVAE extends CF\_VAE by incorporating the DUVAE \citep{Shen2021RegularizingVA}, which improves the diversity of the latent space by applying batch normalization (BN) \citep{ioffe2015batch} and dropout \citep{srivastava2014dropout} to the encoder output of VAE. Similarly to FastDCFlow, a temperature parameter is introduced to VAEs for test input. For comparative reference, we include the evaluation metrics of the input-based methods, namely GeCo \citep{schleich2021geco} and MOC \citep{dandl2020multi}. As these methods are based on GA and differ in parameter types and learning processes, they are not included in our primary comparison set.\\ 

\noindent
\textbf{Parameter setting.}
To ensure a fair assessment, all models applied a consistent ML model based on their encoding methods and the weight of proximity loss is $\bm{w}=\bm{1}$. DiCE, which does not require pretraining, was set for 10 iterations. CeFlow and Model-based methods were subjected to 10 pretraining epochs with a batch size of 64. CeFlow and FastDCFlow assumed a mixed Gaussian distribution \citep{izmailov2020semi, duong2023ceflow}, utilizing RealNVP \citep{dinh2016density} for invertible transformations. Regarding CeFlow, the number of iterations for generating CFs was set to 100. 
For GA-based GeCo and MOC, 10 generations with 1,000 individuals each were established. GeCo randomly generated initial entities, with the next generation comprising the top 100 and 900 offspring produced from a uniform crossover (40\% from Parent 1, 40\% from Parent 2, and a mutation rate 20\%. MOC employed the NSGA2 algorithm \citep{996017}, initiating entities evenly across dimensions using Latin hypercube sampling (LHS) \citep{mckay2000comparison} and populating subsequent generations through uniform crossover with a 50\% crossover rate and a polynomial mutation of the distribution index 20. To determine the effectiveness of TE using FastDCFlow, $N^{tes}=500$ with $M=1000$ were set. In the comparative experiments, when comparing only model-based methods, we set $N^{tes}=500$ with $M=1000$. 
M=1000. For comparisons that involve both model-based and input-based methods, we configure $N^{tes}=100$ with $M=100$. This adjustment is made due to the higher generation costs of input-based methods compared to model-based methods. The temperature parameter $t$ was fixed at $1.0$ in FastDCFlow and VAEs, and the hyperparameter $\lambda$ was determined using a grid search set to 0.01.

\section{Results and analysis}
\label{sec:results_and_analysis}
\begin{figure}[t]
  \begin{center}
  \includegraphics[width=1.0\columnwidth]{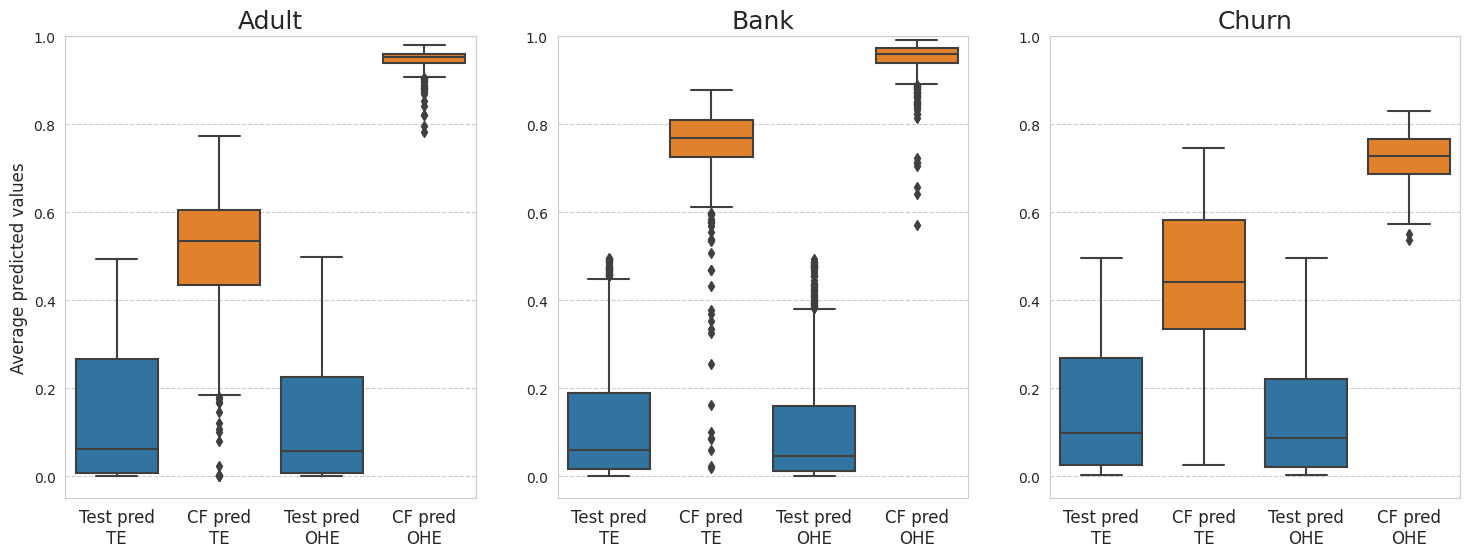}
  \caption{The boxplot showcases $\hat{y}_i^{tes}$ and $\hat{y}_i^{cf}$ values from FastDCFlow trained using TE and OHE across the Adult, Bank, and Churn datasets. In each graph, four boxplots, from left to right, represent: $\hat{y}_i^{tes}$ with TE (blue), $\hat{y}_i^{cf}$ with TE (orange), $\hat{y}_i^{tes}$ with OHE (blue), and $\hat{y}_i^{cf}$ with OHE (orange).}
  \label{fig:boxplot_target_vs_onehot}
  \end{center}
\end{figure}
\begin{table}[t]
\centering
\small
\caption{Evaluation between OHE and TE.}
\begin{tabular}{c|c|c|c}
\hline
\multicolumn{2}{c|}{Dataset} & $\hat{y}^{tes}$ & $\hat{y}^{cf}$ \\ \hline
\multirow{2}{*}{Adult} & OHE & 0.13 & $0.95\pm0.02$ \\
& TE & 0.14 & $0.51\pm \textbf{0.14}$ \\ \hline
\multirow{2}{*}{Bank} & OHE & 0.11 & $0.95\pm0.05$ \\
& TE & 0.13 & $0.75\pm \textbf{0.11}$ \\ \hline
\multirow{2}{*}{Churn} & OHE & 0.14 & $0.72\pm0.05$ \\
& TE & 0.16 & $0.45\pm\textbf{0.16}$ \\ \hline
\end{tabular}
\label{tab:encoding_result}
\end{table}
\subsection{Effectiveness of TE}
\begin{table*}[t]
\centering
\caption{Evaluation results compared with model-based methods at $N^{tes}=500$, $M=1000$. Excluding RT, higher values indicate better performance. The best is marked with a bold.}
\begin{tabular}{c|l|ccccc}
\hline
\multicolumn{2}{c|}{Dataset and model} & ID & OD & P & V & RT (s) \\ \hline
\multirow{4}{*}{Adult} & CF\_VAE & $-0.98_{\pm0.00}$ & $-1.0_{\pm0.00}$ & $-3.2_{\pm0.91}$ & $\bm{1.0_{\pm0.00}}$ & $\bm{2.0\times10^{-3}}$ \\
& CF\_CVAE & $-0.97_{\pm0.00}$ & $-1.0_{\pm0.00}$ & $-3.1_{\pm0.93}$ & $\bm{1.0_{\pm0.00}}$ & $2.8\times10^{-3}$ \\
& CF\_DUVAE & $-0.99_{\pm0.00}$ & ${-1.0_{\pm0.00}}$ & $-3.2_{\pm0.94}$ & $\bm{1.0_{\pm0.00}}$ & $2.2\times10^{-3}$ \\
& FastDCFlow & $\mathbf{-0.78_{\pm0.08}}$ & $\mathbf{-0.33_{\pm0.34}}$ & $\mathbf{-2.4_{\pm0.51}}$ & $0.99_{\pm0.05}$ & $1.5\times10^{-2}$ \\
\hline
\multirow{4}{*}{Bank} & CF\_VAE & $-0.97_{\pm0.00}$ & $-1.0_{\pm0.00}$ & $-4.5_{\pm1.52}$ & $\bm{1.0_{\pm0.00}}$ & $\bm{2.1\times10^{-2}}$ \\
& CF\_CVAE & $-0.98_{\pm0.00}$ & $-1.0_{\pm0.00}$ & $-4.5_{\pm1.52}$ & $\bm{1.0_{\pm0.00}}$ & $2.9\times10^{-3}$ \\
& CF\_DUVAE & $-0.96_{\pm0.00}$ & ${-1.0_{\pm0.00}}$ & $-4.5_{\pm1.52}$ & $\bm{1.0_{\pm0.00}}$ & $2.3\times10^{-3}$ \\
& FastDCFlow & $\mathbf{-0.66_{\pm0.11}}$ & $\mathbf{-0.14_{\pm0.29}}$ & $\mathbf{-2.8_{\pm0.18}}$ & $0.97_{\pm0.04}$ & $7.6\times10^{-3}$ \\
\hline
\multirow{4}{*}{Churn} & CF\_VAE & $-0.97_{\pm0.00}$ & $-1.0_{\pm0.00}$ & $-5.0_{\pm1.38}$ & $0.89_{\pm0.29}$ & $\bm{2.0\times10^{-2}}$ \\
& CF\_CVAE & $-0.97_{\pm0.00}$ & $-1.0_{\pm0.00}$ & $-4.9_{\pm1.33}$ & $0.83_{\pm0.36}$ & $2.8\times10^{-3}$ \\
& CF\_DUVAE & $-0.98_{\pm0.00}$ & ${-1.0_{\pm0.00}}$ & $-5.0_{\pm1.40}$ & $0.84_{\pm0.35}$ & $2.2\times10^{-3}$ \\
& FastDCFlow & $\mathbf{-0.71_{\pm0.09}}$ & $\mathbf{-0.17_{\pm0.32}}$ & $\mathbf{-3.5_{\pm0.56}}$ & $\bm{0.93_{\pm0.09}}$ & $8.7\times10^{-3}$ \\
\hline
\end{tabular}
\label{tab:eval_result_large}
\end{table*}
\begin{table*}[t]
\centering
\caption{Evaluation results compared with input and model-based methods at $N^{tes}=100$, $M=100$.}
\begin{tabular}{c|l|ccccc}
\hline
\multicolumn{2}{c|}{Dataset and model} & ID & OD & P & V & RT (s) \\ \hline
\multirow{8}{*}{Adult} & DiCE & $\bm{-0.26_{\pm0.17}}$ & $-0.92_{\pm0.17}$ & $-4.8_{\pm0.71}$ & $0.80_{\pm0.21}$ & $15$ \\
& CeFlow & $-1.0_{\pm0.00}$ & - & - & - & $6.1$ \\
& CF\_VAE & $-0.98_{\pm0.00}$ & $-1.0_{\pm0.00}$ & $-3.1_{\pm0.86}$ & $\bm{1.0_{\pm0.00}}$ & $\bm{1.4\times10^{-3}}$ \\
& CF\_CVAE & $-0.97_{\pm0.01}$ & $-1.0_{\pm0.00}$ & $-3.0_{\pm0.87}$ & $\bm{1.0_{\pm0.00}}$ & $1.9\times10^{-3}$ \\
& CF\_DUVAE & $-0.99_{\pm0.00}$ & $-1.0_{\pm0.00}$ & $-3.1_{\pm0.87}$ & $\bm{1.0_{\pm0.00}}$ & $1.6\times10^{-3}$ \\
& FastDCFlow & $-0.77_{\pm0.08}$ & $\mathbf{-0.34_{\pm0.33}}$ & $\mathbf{-2.4_{\pm0.50}}$ & $0.99_{\pm0.04}$ & $7.7\times10^{-3}$ \\
\cline{2-7}
& GeCo & $-0.68_{\pm0.17}$ & $-0.016_{\pm0.38}$ & $-1.6_{\pm0.17}$ & $0.87_{\pm0.18}$ & $25$ \\
& MOC & $-0.60_{\pm0.16}$ & $-0.70_{\pm0.20}$ & $-3.0_{\pm0.39}$ & $0.98_{\pm0.02}$ & $30$ \\
\hline
\multirow{8}{*}{Bank} & DiCE & $-0.72_{\pm0.02}$ & $-0.99_{\pm0.00}$ & $-25_{\pm0.84}$ & $0.98_{\pm0.02}$ & $1.4\times10^2$ \\
& CeFlow & $-1.0_{\pm0.00}$ & - & - & - & $5.3$ \\
& CF\_VAE & $-0.98_{\pm0.00}$ & $-1.0_{\pm0.00}$ & $-4.5_{\pm1.57}$ & $\bm{1.0_{\pm0.00}}$ & $\bm{1.6\times10^{-3}}$ \\
& CF\_CVAE & $-0.98_{\pm0.00}$ & $-1.0_{\pm0.00}$ & $-4.5_{\pm1.57}$ & $\bm{1.0_{\pm0.00}}$ & $2.1\times10^{-3}$ \\
& CF\_DUVAE & $-0.96_{\pm0.00}$ & $-1.0_{\pm0.00}$ & $-4.5_{\pm1.57}$ & $\bm{1.0_{\pm0.00}}$ & $1.7\times10^{-3}$ \\
& FastDCFlow & $\mathbf{-0.67_{\pm0.10}}$ & $\mathbf{-0.13_{\pm0.29}}$ & $\bm{-2.8_{\pm0.20}}$ & $0.97_{\pm0.04}$ & $8.7\times10^{-3}$ \\
\cline{2-7}
& GeCo & $-0.76_{\pm0.10}$ & $-0.080_{\pm0.29}$ & $-2.0_{\pm0.16}$ & $0.98_{\pm0.05}$ & $39$ \\
& MOC & $-0.73_{\pm0.06}$ & $-0.90_{\pm0.11}$ & $-11_{\pm1.17}$ & $1.0_{\pm0.00}$ & $29$ \\
\hline
\multirow{8}{*}{Churn} & DiCE & $\mathbf{-0.26_{\pm0.02}}$ & $-0.97_{\pm0.01}$ & $-5.8_{\pm0.33}$ & $0.77_{\pm0.28}$ & $1.4\times10^2$ \\
& CeFlow & $-1.0_{\pm0.00}$ & - & - & - & $4.9$ \\
& CF\_VAE & $-0.97_{\pm0.00}$ & $-1.0_{\pm0.00}$ & $-5.3_{\pm1.35}$ & $0.93_{\pm0.23}$ & $\bm{1.5\times10^{-3}}$ \\
& CF\_CVAE & $-0.97_{\pm0.00}$ & $-1.0_{\pm0.00}$ & $-5.2_{\pm1.30}$ & $0.83_{\pm0.36}$ & $1.9\times10^{-3}$ \\
& CF\_DUVAE & $-0.98_{\pm0.00}$ & $-1.0_{\pm0.00}$ & $-5.3_{\pm1.36}$ & $0.85_{\pm0.33}$ & $1.6\times10^{-3}$ \\
& FastDCFlow & $-0.72_{\pm0.09}$ & $\mathbf{-0.20_{\pm0.34}}$ & $\mathbf{-3.6_{\pm0.58}}$ & $\mathbf{0.94_{\pm0.07}}$ & $9.6\times10^{-3}$ \\
\cline{2-7}
& GeCo & $-0.92_{\pm0.03}$ & $-0.041_{\pm0.39}$ & $-5.1_{\pm1.75}$ & $0.62_{\pm0.45}$ & $43$ \\
& MOC & $-0.78_{\pm0.06}$ & $-0.48_{\pm0.25}$ & $-6.0_{\pm1.2}$ & $0.96_{\pm0.08}$ & $30$ \\ \hline
\end{tabular}
\label{tab:eval_result_small}
\end{table*}
Figure \ref{fig:boxplot_target_vs_onehot} shows the difference in the predicted values between each test input and its counterfactuals, and the evaluation results are presented in Table \ref{tab:encoding_result}. 
For every dataset, the OHE model consistently improved the predicted target values. For the Adult and Bank datasets, $\hat{y}_i^{cf}$ had an upper limit of 1.0, whereas Churn showed better results with OHE. However, the standard deviation values for OHE were smaller across datasets compared to those for TE. This suggests that OHE consistently generated counterfactuals irrespective of the test input features because categorical variables have a uniform perturbation cost in OHE. In contrast, TE, which transforms categorical variables into continuous ones, considered varying costs and better reflected the traits of each test input, generating more diverse predicted values.

\subsection{Overall Performance}
Table \ref{tab:eval_result_large} presents the evaluation results for the comparison of model-based methods. FastDCFlow demonstrates superior performance in ID, OD, and P metrics compared to existing VAE-based methods. Models rooted in the VAE, including CF\_VAE, CF\_CVAE, and CF\_DUVAE, consistently generated similar CFs irrespective of the test input variation. Particularly, the performance in OD being the minimum at -1.0 indicates that the model is not considering the differences between inputs at all. This can be attributed to their inherent leanings toward minimizing the KL divergence over optimizing the predictive accuracy, which is particularly evident in CF\_VAE. Whereas, although CF\_DUVAE sought latent space diversity, it fell short of with data rich in categorical variables, suggesting that the VAEs Gaussian distribution assumption in the latent space might not be suitable for intricate tabular data. 

Meanwhile, FastDCFlow has successfully improved performance in ID and OD while maintaining high levels in P and V. In particular, the significant improvement in OD suggests that the model accurately accounts for variations in input. The highest values in P indicate that the model has learned a latent space which preserves proximity, even as inputs change. Regarding V, FastDCFlow is slightly outperformed by VAEs in the Adult and Bank datasets, yet the difference is marginal, and over 90\% of the generated CFs show improved prediction values. As for RT, there is little variation in performance across models, with all completing execution in less than 1.0s per CF, indicating efficiency.

The evaluation results comparing input and model-based methods are shown in Table \ref{tab:eval_result_small}. 
In the comparison with VAEs, the results are almost identical to those in Table \ref{tab:eval_result_small}, indicating that performance is not dependent on the number of test inputs or the number of generated CFs. In the context of input-based methods, the focus on DiCE reveals a tendency to achieve higher ID performance compared to other models, indicating greater diversity within the generated CFs. This is likely due to DiCE's use of a determinant-based explicit diversity loss, resulting in CFs that differ significantly in direction. However, in terms of OD, P, V and RT, overall performance is inferior, suggesting an imbalance with metrics other than ID. For CeFlow, the ID value of -1.0 indicates a complete lack of consideration for CF diversity. This is attributed to the fact that CeFlow is designed to generate a single CF, leading to identical outcomes for all optimization results of perturbations.

Focusing on FastDCFlow, it demonstrates competitive performance across all metrics compared to other models, particularly excelling in OD and P. However, when considering the results of GeCo and MOC, it is observed that in the Adult dataset, GeCo outperforms overall, but in other datasets, FastDCFlow exhibits equivalent or superior performance. Furthermore, the RT results indicate that model-based methods are faster than input-based methods, demonstrating the efficiency of optimization using the latent space. Consequently, FastDCFlow is capable of generating CFs that achieve a balance of metrics in a shorter computation time.

\subsection{The effect of CF parameters}
\begin{figure}[t]
    \begin{center}
    \includegraphics[width=1.0\columnwidth]{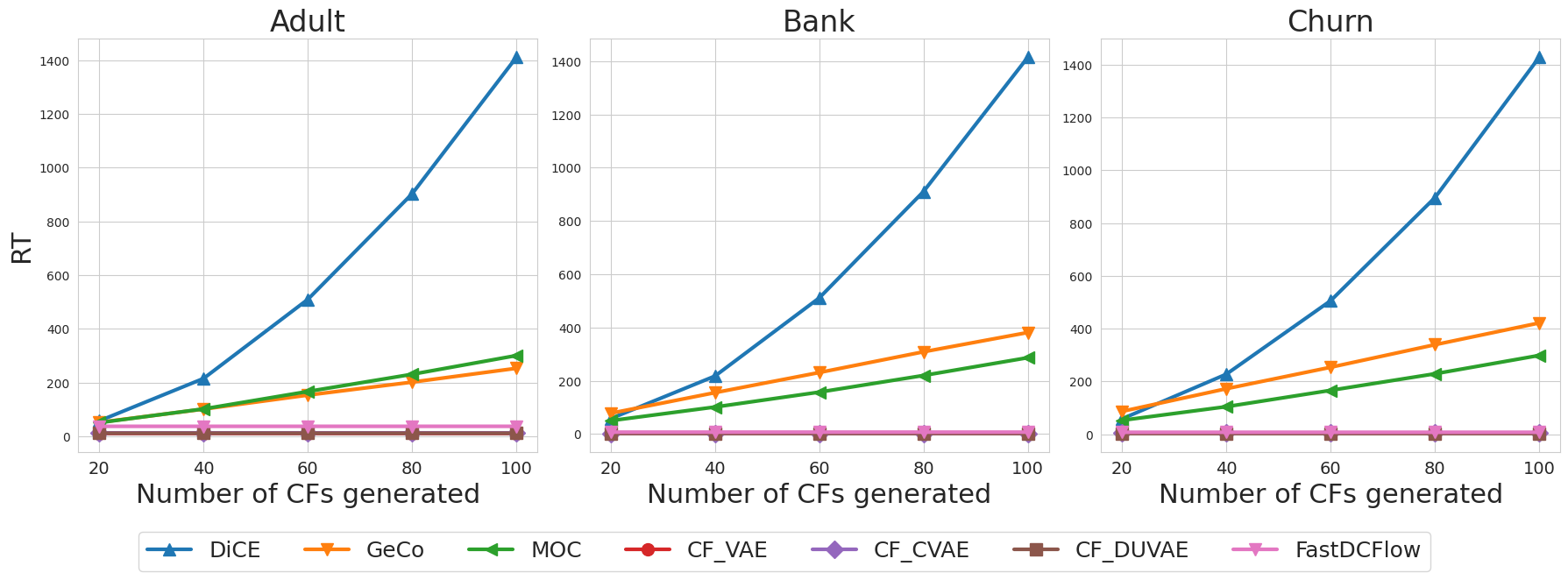}
    \caption{Relationship between $M$ and RT for each model.}
    \label{fig:runtime}
    \end{center}
\end{figure}
\begin{figure}[t]
    \begin{center}
    \includegraphics[width=1.0\columnwidth]{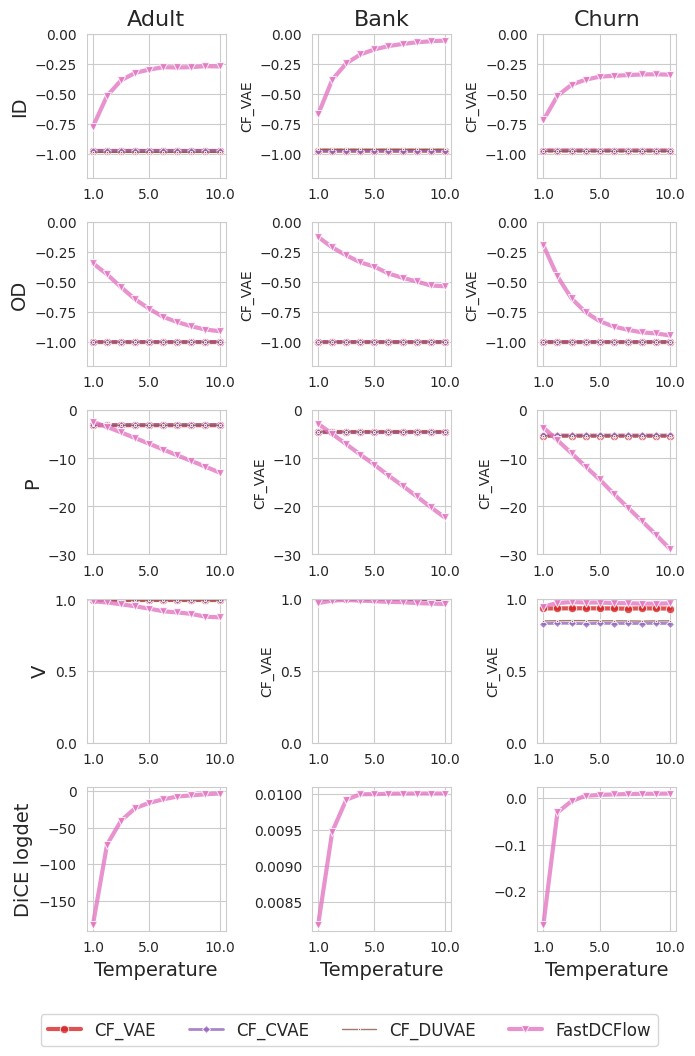}
    \caption{Changes in evaluation metrics with adjustments to the temperature parameter in model-based methods. Each line represents an evaluation metric (including DiCE diversity loss), and each column corresponds to a dataset. VAE-based methods produce similar values in all evaluations, leading to nearly overlapping plots.}
    \label{fig:temperature}
    \end{center}
\end{figure}
\noindent
\textbf{Sample size effect.} Figure \ref{fig:runtime} presents a comparison of the relationship between $M$ and RT ($N^{tes}=10$ and $M=20, 40, \cdots, 100$). Model-based methods, regardless of the number of dimensions in the data, can significantly reduce the RT. These models, excluding pretraining time, generate CFs through latent space mapping and become more practical with increase in $M$. The RT of DiCE increases exponentially with the number of generations, owing to the computational complexity of $\mathcal{O}(KM^2 + M^3)$ required for the diversity loss term in datasets with dimension number $K$.\\

\noindent
\textbf{Temperature effect.} Figure \ref{fig:temperature} shows the trends of ID, OD, P, and V when varying the temperature parameter of model-based methods ($N^{tes}=100$ and $M=100$). In the case of FastDCFlow, an increase in temperature results in a rise in ID and a decrease in P, highlighting a trade-off between internal diversity and proximity. Furthermore, a contrast between ID and OD is observed, implying that higher temperatures lead to increased randomness in counterfactual samples and enhanced similarity among their means. V is defined as the proportion of improved predictions for test inputs; therefore, it is possible to enhance diversity without deteriorating its value. On the contrary, VAE-based methods showed no variation in the evaluation metrics regardless of temperature changes, suggesting that these models do not adapt to the characteristics of the test inputs. Furthermore, to investigate the relationship between DiCE's explicit diversity loss and FastDCFlow temperature, the changes in the logdet value of the diversity loss 
are also shown in line 5. The results confirm that the logdet value increases with temperature changes and reaches a steady state when the temperature is sufficiently high.

\subsection{The effect of loss functions}
\begin{table*}[t]
\centering
\caption{The ablation study of FastDCFlow.}
\label{tab:fastdcflow_ablation}
\begin{tabular}{l|l|cccc}
\hline
\multicolumn{2}{c|}{Dataset and method} & \multicolumn{1}{c}{ID} & \multicolumn{1}{c}{OD} & \multicolumn{1}{c}{P} & \multicolumn{1}{c}{V} \\\hline
\multirow{3}{*}{Adult} & $\mathcal{L}_{nll}$ & $-0.46_{\pm0.14}$ & $-0.031_{\pm0.39}$ & $-2.7_{\pm0.02}$ & $0.42_{\pm0.05}$ \\
& $\mathcal{L}_{nll}+\mathcal{L}_{y}$ & $-0.79_{\pm0.13}$ & $-0.94_{\pm0.08}$ & $-18_{\pm3.86}$ & $0.98_{\pm0.05}$ \\
& $\mathcal{L}_{nll}+\mathcal{L}_{y}+\mathcal{L}_{wprox}$ & $-0.78_{\pm0.08}$ & $-0.33_{\pm0.34}$ & $-2.4_{\pm0.51}$ & $0.99_{\pm0.05}$ \\\hline
\multirow{3}{*}{Bank} & $\mathcal{L}_{nll}$ & $-0.43_{\pm0.14}$ & $-0.091_{\pm0.29}$ & $-3.9_{\pm0.02}$ & $0.55_{\pm0.02}$ \\
& $\mathcal{L}_{nll}+\mathcal{L}_{y}$ & $-0.098_{\pm0.04}$ & $-0.47_{\pm0.23}$ & $-19_{\pm0.35}$ & $1.0_{\pm0.01}$ \\
& $\mathcal{L}_{nll}+\mathcal{L}_{y}+\mathcal{L}_{wprox}$ & $-0.66_{\pm0.11}$ & $-0.14_{\pm0.29}$ & $-2.8_{\pm0.18}$ & $0.97_{\pm0.04}$ \\\hline
\multirow{3}{*}{Churn} & $\mathcal{L}_{nll}$ & $-0.51_{\pm0.08}$ & $-0.026_{\pm0.36}$ & $-4.1_{\pm0.02}$ & $0.35_{\pm0.05}$ \\
& $\mathcal{L}_{nll}+\mathcal{L}_{y}$ & $-0.86_{\pm0.06}$ & $-0.97_{\pm0.02}$ & $-23_{\pm3.37}$ & $1.0_{\pm0.01}$ \\
& $\mathcal{L}_{nll}+\mathcal{L}_{y}+\mathcal{L}_{wprox}$ & $-0.71_{\pm0.09}$ & $-0.17_{\pm0.32}$ & $-3.5_{\pm0.56}$ & $0.93_{\pm0.09}$ \\\hline
\end{tabular}
\end{table*}
To evaluate the impact of adding different loss functions, we performed an ablation study on FastDCFlow. The study compared three models: one with only the likelihood loss denoted $\mathcal{L}_{nll}$, another with the combination of likelihood and validity losses represented as $\mathcal{L}_{nll} + \mathcal{L}_{y}$, and the proposed FastDCFlow method, which includes likelihood, validity and weighted proximity losses, indicated as $\mathcal{L}_{nll} + \mathcal{L}_{y} + \mathcal{L}_{wprox}$. We present the results of these comparisons for the Adult, Bank, and Churn datasets at $N^{tes}=500$ and $M=1000$ in Table \ref{tab:fastdcflow_ablation}.

The model with only likelihood loss shows good results in ID, OD, and P, but a deterioration in V performance. Likelihood loss alone merely approximates the observed distribution of the data and fails to consider the constraints of V that counterfactuals should satisfy. Adding validity loss significantly improves V performance. However, since it learns a latent space disregarding the input features, the performance in OD and P is poor. FastDCFlow, with the addition of proximity loss, improves performance in OD and P while maintaining V. Furthermore, as ID performance is on par with existing methods shown in the table \ref{tab:eval_result_large} and \ref{tab:eval_result_small}, the proposed approach exhibits the most balanced performance overall.

\section{Conclusion}
\label{sec:conclusion}
This study applied a TE transformation that considered the perturbation of categorical variables in CE, thus tackling the problem of CF's predicted values tending to the upper limit. In addition, we introduced a method to learn the latent space in alignment with the CF constraints by utilizing a normalizing flow termed FastDCFlow. This latent space, which was derived from the training data, offered efficient generation for any test input. Our experimental findings confirmed the proficient balance that this approach maintained among the evaluation metrics, making it the best in terms of many aspects. FastDCFlow's superior performance was largely due to its advances in diversity and speed over input-based methods. Moreover, in contrast to VAE-based methods, FastDCFlow captured a precise depiction of the training data, enabling CF generation through the inverse transformation of proximate points. 

A limitation of this work is fourfold. First, when applying TE with certain categories that feature rarely seen levels or are data-deprived, there is a looming threat of overfitting, owing to stark fluctuations in the target variable. Second, given that decision-makers are human, it is impossible to ignore user feedback and biases originating from the domain. Although FastDCFlow offers new possibilities for utilizing model-based methods, qualitative evaluation of the generated CEs and detection of biases remain as future work. Thirdly, as both input-based and model-based methods require ML training, sparse high-dimensional data can hinder the utility of CEs. Fourth, the applicability of ML is limited to differentiable models. Although decision tree-based models are commonly used with tabular data, they are not applicable in gradient-based models.

In subsequent phases of this study, a deeper exploration is imperative to understand the underlying mechanisms of how the model depicts intricate tabular data distributions.

\bibliographystyle{./IEEEtran}
\bibliography{./IEEEfull}

\begin{table*}[t]
    \centering
    \caption{The results of domain adaptation.}
    \label{tab:domain_constraints}
    \begin{tabular}{l|cccccc}
    \hline
    \multicolumn{1}{c|}{Method} & \multicolumn{1}{c}{FA} & \multicolumn{1}{c}{MA} & \multicolumn{1}{c}{ID} & \multicolumn{1}{c}{OD} & \multicolumn{1}{c}{P} & \multicolumn{1}{c}{V} \\\hline
    No constraints & $0.97_{\pm0.01}$ & $0.88_{\pm0.21}$ & $\mathbf{-0.78_{\pm0.08}}$ & $\mathbf{-0.33_{\pm0.34}}$ & $-2.4_{\pm0.51}$ & $\mathbf{0.99_{\pm0.05}}$ \\
    Domain constraints & $\mathbf{0.99_{\pm0.01}}$ & $\mathbf{0.91_{\pm0.12}}$ & $-0.80_{\pm0.09}$ & $-0.34_{\pm0.33}$ & $\mathbf{-2.3_{\pm0.50}}$ & $0.97_{\pm0.10}$ \\\hline
    \end{tabular}
\end{table*}
\begin{table*}[t]
\centering
\caption{Top 5 CFs by Likelihood for FastDCFlow with and without domain constraints. Items that meet the constraints are highlighted in bold.}
\label{tab:cfs_likelihood}
\begin{tabular}{llllllrrr}
\hline
\textbf{Race} & \textbf{Gender} & \textbf{Workclass} & \textbf{Education} & \textbf{Marital Status} & \textbf{Occupation} & \textbf{Age} & \textbf{Hours per Week} & \textbf{Income} \\ \hline
\multicolumn{9}{c}{Test input} \\ \hline
White & Male & Private & School & Married & Blue-Collar & 47 & 40 & 0.17 \\
\hline
\multicolumn{9}{c}{CFs (no constraints)} \\ \hline
\textbf{White} & \textbf{Male} & Private & HS-grad & Married & Sales & \textbf{48} & 45 & 0.58 \\
\textbf{White} & \textbf{Male} & Private & HS-grad & Married & Sales & 46 & 48 & 0.53 \\
\textbf{White} & Female & Private & HS-grad & Married & White-Collar & 47 & 51 & 0.61 \\
\textbf{White} & \textbf{Male} & Private & Some-college & Married & Sales & 47 & 51 & 0.64 \\
\textbf{White} & \textbf{Male} & Private & Some-college & Married & White-Collar & \textbf{53} & 44 & 0.66 \\ \hline
\multicolumn{9}{c}{CFs (domain constraints)} \\ \hline
\textbf{White} & \textbf{Male} & Private & Bachelors & Married & Sales & \textbf{54} & 56 & 0.75 \\
\textbf{White} & \textbf{Male} & Private & Assoc & Married & Sales & \textbf{52} & 52 & 0.69 \\
\textbf{White} & \textbf{Male} & Private & Assoc & Married & Sales & \textbf{54} & 50 & 0.67 \\
\textbf{White} & \textbf{Male} & Private & Assoc & Married & Sales & \textbf{54} & 47 & 0.68 \\
\textbf{White} & \textbf{Male} & Private & Assoc & Married & White-Collar & \textbf{57} & 45 & 0.62 \\ \hline
\end{tabular}
\end{table*}

\appendix

\section*{Applying domain constraints}
\label{adx:applying_domain_constraints}
Depending on the type of dataset and specific domain constraints aligned with the objectives, it is possible to apply counterfactual explanations appropriately. In reality, characteristics such as gender or race cannot be perturbed. Furthermore, age must always satisfy a monotonic increase, making explanations that do not adhere to these constraints infeasible. This chapter discusses the configuration of mutable features and monotonic constraints using FastDCFlow.

Initially, the perturbation constraints on the features can be achieved by adjusting the weights $\bm{w}$ of the proximity loss.
\begin{equation}
    \mathcal{L}_{wprox}(\bm{\theta}, \mathcal{D}) = \sum_{i=1}^{N^{tra}}\| \bm{w}*(\mathbf{x}_{i}^{tra} - \mathbf{x}_{i}^{cf}) \|_2^{2}.
\end{equation}
Specifically, increasing the weights for features that should not be perturbed effectively imposes stronger penalties during perturbation. For the monotonic constraints, the following hinge loss function is added:
\begin{equation}
    \mathcal{L}_{mon} = \frac{1}{|\mathcal{D}^{mon}|N^{tra}}\sum_{i=1}^{N^{tra}}\sum_{d\in D^{mon}}\max(x_{id}^{tra}-x_{id}^{cf}, 0),
\end{equation}
where $D^{mon}$ is the set of features that should be monotonically increased. 

In the context of the Adult dataset, we compared the performance before and after the addition of domain constraints. For the evaluation, we applied fix accuracy (FA) and monotonicity accuracy (MA) metrics to the features with imposed constraints and also use P, V, ID, and OD used in Table \ref{tab:eval_result_large} for the overall evaluation.\\
\begin{equation}
    \text{FA} = \frac{1}{N^{tes}}\sum_{i=1}^{N^{tes}}\frac{1}{|\mathcal{D}^{fix}|M}\sum_{j=1}^{M}\sum_{d \in \mathcal{D}^{fix}} I\left(x_{ijd}^{cf} =x_{id}^{tes}\right),
\end{equation}
\begin{equation}
    \text{MA} = \frac{1}{N^{tes}}\sum_{i=1}^{N^{tes}}\frac{1}{|\mathcal{D}^{mon}|M}\sum_{j=1}^{M}\sum_{d \in \mathcal{D}^{mon}} I\left(x_{ijd}^{cf} > x_{id}^{tes}\right),
\end{equation}
where $\mathcal{D}^{fix}$ is the sets of features with fixed constraints. FA and MA are metrics that evaluate the proportion of CFs that satisfy the fixed and monotonic constraints, respectively. These metrics are evaluated by returning the input data and CFs to their original scale and category before evaluation.

For the features that should remain fixed, ``gender" and ``race" are selected, with their respective weight coefficients set to 3.0, while setting the coefficients for all other features to 1.0. Additionally, ``age" is designated as a feature that must satisfy the monotonic increase constraint. The results are shown in Table \ref{tab:domain_constraints} for $N^{tes}=500$ and $M=1000$. The results show that the addition of domain constraints outperforms the model without constraints in the FA, MA, and P metrics. This suggests that the addition of domain constraints can be effective in generating CFs that are more appropriate for the domain. However, it is important to note that the addition of domain constraints can lead to a decrease in diversity. Therefore, it is necessary to consider the trade-off between diversity and domain constraints when applying domain constraints.

Table \ref{tab:cfs_likelihood} presents the top 5 CFs with the highest likelihood values generated for a specific test input. Observing FastDCFlow without domain constraints, we can confirm the existence of CFs that do not satisfy constraints in features such as ``gender" and ``age". On the other hand, FastDCFlow with constraints shows that all samples meet the constraints, indicating a tendency to consider domain-specific characteristics. These results suggest that by appropriately setting domain constraints according to the data and users, more flexible CFs can be generated.

\section*{Model architectures}
\label{adx:model_architectures}
The ML model for predicting the target variable is constructed using a three-layer binary classification neural network. ReLU activation functions and dropout (with 0.5) are applied between each layer. The final output is obtained as a probability in the range $[0, 1]$ from a fully connected layer with 64 units, passed through a sigmoid function. The gradient descent optimization algorithm used is Adam \citep{kingma2014adam}, with a fixed learning rate of $10^{-3}$. The encoders for CF\_VAE, CF\_CVAE, and CF\_DUVAE are multi-layer neural networks that incorporate BN, dropout (with 0.1), and the ReLU. VAE and DUVAE have three layers, while CVAE, due to the added dimensionality of labeling categorical variables, is constructed with four layers. For CF\_DUVAE, when mapping to the latent space, BN is applied to the mean, and dropout (with 0.5) is applied to the variance. Regarding the RealNVP used in FastDCFlow, there are 3 coupling layers, and each coupling layer consists of 6 intermediate layers containing ReLU and dropout. 

All codes are executed in Python 3.9, equipped with the PyTorch library. The machine specifications used for the experiments are Debian GNU/Linux 11 (bullseye), Intel Core i9 3.30GHz CPU, and 128GB of memory.
\end{document}